\documentclass[journal]{IEEEtran}
\usepackage{amsmath}
\usepackage{algorithm}
\usepackage[noend]{algpseudocode}
\usepackage{graphicx}
\usepackage{array}
\usepackage{float}
\usepackage{capt-of}
\graphicspath{ {/} }

\makeatletter
\def\BState{\State\hskip-\ALG@thistlm}
\makeatother
\ifCLASSINFOpdf
\else
\fi
\begin{document}

%
\title{BACKGROUND MODELING USING OCTREE COLOR QUANTIZATION}
%
%
%

\author{AV Aditya Sastry,B.Tech, GITAM University}
\maketitle

\begin{abstract}
By assuming that the most frequently occuring color in a video or a region of a video belongs to the background of the image, I propose a new algorithm for detecting foreground objects in a video. The process of detecting the foreground objects is complicated because of the fact that there may be swaying tree’s, objects of the background being moved around or lighting changes in the video. To deal with such complexities many have come up with solutions which heavily rely on expensive floating point operations. In this paper I used a data structure called Octree which is implemented only using binary operations. Traditionally octrees were used for color quantization but here in this paper I used it as a data structure to store the most frequently occuring colors in a video as well. For each of the starting few video frames, I constructed a Octree using all the colors of that frame. Next I pruned all the trees by removing nodes below a certain height and gave the leaf nodes a color which is dependant on the topological path from that node to its parent. Hence any two leaf nodes in two different octrees with the same topological path from themselves to the root will represent the same color. Next I merged all these individual trees into a single tree retaining only those nodes whose topological path to itself from the root is most common among all the trees. The colors represented by the leaf nodes in the resultant tree will be the most frequently occuring colors in the starting video frames of the video. Hence any color of an incomming frame that is not close to any of the colors represented by the leaf node of the merged tree can be regarded as belonging to a foreground object.\\ 

As an Octree is constructed using only binary operations, it is very fast compared to other leading algorithms.
\end{abstract}

\begin{IEEEkeywords}
Background modelling, Octree color quantization, moving object detection
\end{IEEEkeywords}

%
\IEEEpeerreviewmaketitle

\section{Introduction}
%
%
%
%
\IEEEPARstart {D}ue to the advancement of technology and heavy reduction in costs of FPGA development, traditional video surveillance equipment has evolved from merely capturing video from a static vantage point and transmitting it to performing some preliminary analysis such as extracting objects of interest from the scene and then transmitting it.\\    

	For most of the analysis that is done on a surveillance video, the first crucial step is to detect theforeground pixels of the video. This is done by first constructing a background model and matching each individual frame with this model. Any pixel of a frame which deviates significantly from this model is marked as a foreground pixel in that image. \\        

	Several techniques for construction of this background model have been proposed. The most trivial of this techniques is to average the first 20 or so frames and use the averaged frame as the background model. This adequately takes care of the lighting changes but this technique fails when the background of the scene contains swaying leaves, clocks or part of the background is constantly moved around the scene etc.\\

	Wren et al,[1] tried to solve this problem by fitting a Gaussian distribution for all the colors occuring at each pixel location of a frame using the past N frames. Any pixel value from the input frame which is sufficiently close to the mean of the Gaussian distribution for that pixel location is marked as a background pixel and the Gaussian distribution is updated. This technique fails when the histogram of pixel values at any given location is bimodal or has multiple peeks. \\ 

	Stauffer et al,[2] attacked the short comings of the previous work by using multiple(about 3 to 5) Gaussian distributions per pixel locations. Each input frame pixel is matched with a corresponding Gaussians at that location. Any pixel which is in the range of 2.5σ of the mean of a Gaussian for that pixel location is marked as a background pixel and that Gaussian’s parameters are updated. The problem with this technique is its heavy usage of floating point operations and its huge memory requirement. Also the error rate is very high. \\ 

	The wallflower principles and practises [3] a pioneering work by Microsoft research team, not only came up with a background modelling technique but also surveyed all the recent background modelling techniques and listed out all the common problems that a background modelling techniques would face. It views the process of constructing the background model at pixel level, segment level and frame level. The pixel level algorithm attacks the problems with leaf swaying and background motion problems. In this work I provide a better alternative to their pixel level algorithm. The frame level and the segment level techniques listed out in the wallflower paper could be used with this paper as well.\\ 

    In this paper I tackle the problem by constructing an a octree[5] which is discussed in section 2. The octree is constructed by iterating over each individual pixel location of each individual frame of the training video. The colors found are quantized and form a leaf node in the octree. Each pixel of the incoming frame is quantized similarly and checked to see if a corresponding leaf node exists in the already constructed octree. This is shown in section 3. If a node does exists it is marked as a background pixel. In section 4 I show a method of merging multiple octrees a technique to be used when we don't have a trainining video or as the wallflower paper calls it, the bootstrapping problem.\\ 

\section{Octree color quantization}

	An octree is a tree data structure in which each internal node has exactly eight children. Octrees are most often used to partition a three dimensional space by recursively subdividing it into eight octants.  Octrees are also used for color quantization. The octree color quantization algorithm, invented by Gervautz and Purgathofer in 1988, encodes image color data as an octree up to nine levels deep. \\ 

	In computer graphics, color quantization or color image quantization is a process that reduces the number of distinct colors present in an image, usually with the intention that the new image should be as visually similar as possible to the original image. Various methods are available for color quantization such as the median cut algorithm, popularity algorithm. 
Octree[5] is the leading color quantization technique as it executes in $O(N)$ execution time and barely occupies memory of the rder $O(K)$. Here $N$ being the number of pixels in the image and $K$ is a number which is proportional to the number of unique colors present.\\  

\begin{center}
\includegraphics[width=2.5in]{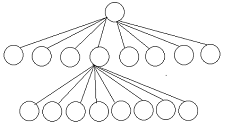}  
\captionof{figure} {Visual representation of a octree. }
\end{center}  

	For a picture whose colors we wish to quantize, we consider each color and store it in a octree using the store color algorithm. Next we prune the octree. By pruning the octree we hope to group together colors which look visually same. Few methods of pruning have been covered in [14] and [15]. In this paper I followed a simple pruning algorithm that is to remove few levels from the bottom of the tree. Since each color at a level N is constructed using N most significant bits of the red, green, blue values of the color, the newly exposed leaf nodes will have N bits of the original color that they represented.One way to assign a new color to these newly exposed leaf nodes is to use their N original bits and fill out the remaining bits with 0's. This way if any two colors had the same parents till the height of pruning will now be reprsented by the same color. This new color could be used to redraw the image. Fig 2 shows the same image redrawn using an octree pruned at different levels.
 
\begin{algorithm}[H]
\caption{Store color}\label{EQUAL SIZED SUBCLUSTERING}
\begin{algorithmic}[1]
\Procedure{Store Color}{ Color, Root, Number of Levels}
\If {$\text {Root is NULL }$} 
\State $\text{root} \gets \text{create node();}$
\EndIf
\State $\text{node temp} \gets \text{root}$
\For{\texttt{$i := \text{MSB  to Number of Levels}$}}
\State $\text{index} \gets \text{Color.red[i] + Color.green[i] + Color.blue[i];}$
 \If {$\text{temp.child[index]} \neq \text{NULL}$}
\State $\text{temp} \gets \text{temp.child[index]}$
\Else
\State $\text{temp.child[index]} \gets \text{create node();}$
\State $\text{temp} \gets \text{temp.child[index]}$
\EndIf
\State $\textbf{End If } $
\EndFor
\State $\textbf{End For}$
\EndProcedure{\textbf{End}}
\end{algorithmic}
\end{algorithm}

\begin{figure}
\centering
\includegraphics[width=2.5in]{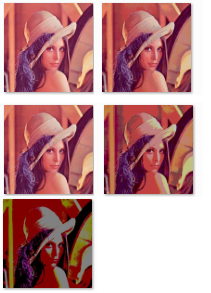}  
\caption{Original image of lena 2.a being quantized using a octree 5 levels deep 2.b, 4 levels deep 2.c, 3 levels deep 2.d, 2 levels deep 2.e}
\end{figure}

\section{Constructing an octree containing only the background colors of an image}

	If we assume that the most frequently occuring color in a video usually belong to the background, we can construct a octree only containing the background colors. We construct an octree for each frame of a video or a group of frames and prune it using the technique presented in the previous section. Each octree obtained this way will hold the quantized representation of all the colors in each frame or a group of frames. By using the merging procedure shown below we could merge together multiple octrees to obtain a single octree which will contain the most frequently occuring colors in a quantized way. This octree could be thought of as containing all the colors of the background.\\ 

	Our assumption that the most frequently occuring colors in a video usually belong to the background is invalidated in the case when a foreground object stays static for a while and suddenly starts to move. Since that object's color has appeared in most frames it would be treated as belonging to the background of the video. This could happen when a person sleeping infront of a camera suddenly starts  to move. To remedy this, we could subdivide a video into multiple regions and obtain a octree for each of these regions. \\ 

\begin{algorithm}
\caption{Merge multiple octrees into a single octree}\label{EQUAL SIZED SUBCLUSTERING}
\begin{algorithmic}[1]
\Procedure{Merge Trees}{ Roots[], Threshold, numberOfTrees, Number of Levels}
\For{\texttt{$i := 1\ to\ 8$}}
\State $\text{count} \gets 0$
\For{\texttt{$j := 1\ to \ numberOfTrees$}}
\If {$\text{roots[j]} \neq \text{NULL}\ and\ \text{roots[j].child[i]} \neq \text{NULL}$}
\State $\text{count} \gets \text{count + 1;}$
\EndIf
\State $\textbf{End if}$
\EndFor
\State $\textbf{End For}$
\If {$\text{count/numberOfTrees} \geq \text{threshold}$}
\State $\text{result.child[i]} \gets \text{New Node();}$
\EndIf
\State $\textbf{End If}$
\EndFor
\State $\textbf{End For}$
\For{\texttt{$i := 1\ to\ 8$}}
\If {$\text{result.child[i]} \neq \text{NULL}$}
\For{\texttt{$j := 1\ to\ numberOfTrees$}}
\State $\text{tempRoot[j]} \gets \text{root[j].child[i];}$
\EndFor
\State $\textbf{End For}$
\State $\text{Result.child[i]} \gets \text{Construct Tree(tempRoot,}$
\Statex $\text{ threshold, numberOfTrees, Number of levels);}$
\EndIf
\State $\textbf{End If}$
\EndFor
\State $\textbf{End For}$
\EndProcedure{\textbf{End}}
\end{algorithmic}
\end{algorithm}

\section{Foreground pixel detection and octree updation. }

	To identify the foreground pixels in a video, we run the procedure shown in the previous section on the first few seconds of the video or the first 100 frames or so of the video. Once we constructed an octree containing the background colors, the process of identifying foreground object locations in a incomming frame is pretty straight forward. For each pixel color of the incoming frame, we use the check color procedure to see if it is present in the octree. If it is not, it is labelled as a foreground pixel.\\ 

	In case we subdivided the video into multiple regions when checking if a pixel belongs to foreground or background, we use the octree of that region in conjugation with the procedure below to ascertain if it is foreground or background. 

\begin{algorithm} [H]
\caption{Check color}\label{EQUAL SIZED SUBCLUSTERING}
\begin{algorithmic}[1]
\Procedure{Check Color}{ Color, Root, Number of Levels}
\If {$\text {Root is NULL }$} 
\State $\textbf{return} \ \text{False}$
\Else
\State $\text{node temp} \gets \text{root}$
\For{\texttt{$i := \text{MSB  to Number of Levels}$}}
\State $\text{index} \gets \text{Color.red[i] + Color.green[i] + Color.blue[i];}$
 \If {$\text{temp.child[index]} \neq \text{NULL}$}
\State $\text{temp} \gets \text{temp.child[index]}$
\Else
\State $\textbf{return} \ \text{False}$
\EndIf
\State $\textbf{End If } $
\EndFor
\State $\textbf{End For}$
\State $\textbf{return} \ \text{True}$
\EndIf
\State $\textbf{End If}$
\EndProcedure{\textbf{End}}
\end{algorithmic}
\end{algorithm}

\section{Experimental Results}
	The following results are obtained on a second generation i5 laptop running at 2.3 GHZ with 4 GB of ram on Windows 7 SP1 and JRE 7 with no updates installed. The datasets for these experimentation is from the wallflower paper. 

\begin{center}
  \includegraphics{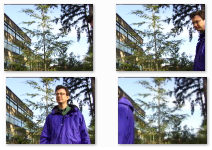}
  \captionof{figure}{A video with waving trees in the background}
\end{center}

\begin{center}
\includegraphics[width=2.5in]{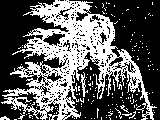}  
\captionof{figure}{Result of motion detection using frame on frame differencing }
\end{center}

\begin{center}
\includegraphics[width=2.5in]{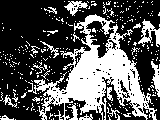}  
\captionof{figure}{Result of running frame averaging}
\end{center}

\begin{center}
\includegraphics[width=2.5in]{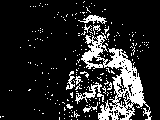}  
\captionof{figure}{Modeling using mixture of Gaussians. }
\end{center}

\begin{center}
\includegraphics[width=2.5in]{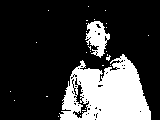}  
\captionof{figure}{Modeling using octree 4 levels deep. }
\end{center}

\begin{center}
\includegraphics[width=2.5in]{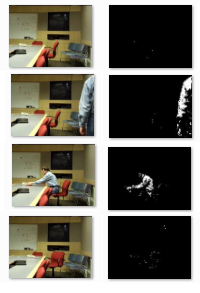}  
\captionof{figure}{Motion detection using an octree 4 levels deep.}
\end{center}

\begin{center}
\includegraphics[width=2.5in]{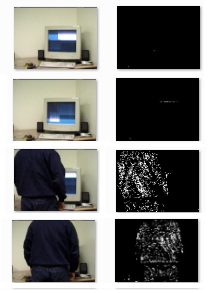}  
\captionof{figure}{Motion detection using an octree  levels deep.}
\end{center}

The $F_0$ scores for different leading algorithms(CB[6], IMOG[8], LBPH[9], STBS[11]) on different datasets is given below in comparision with Octree. The $F_0$ scores are calculated using the formula given below.\\ 

$F_0 = 2\frac{P_0\  R_0}{P_0 + R_0}$\\ 

Where $P_0 = \frac{T_n}{T_n + F_n}$ and $R_0 = \frac{T_n}{T_n + F_p}$\\ 

$T_n, T_p, F_n, F_p$  stand for True negative, True positives, False negatives and False Positives.

\begin{table}[H]
\begin{tabular}{|l|l|l|l|l|l|}
\hline
       & EE    & SW    & IND   & WK    & Avg      \\ \hline
CB     & 0.973 & 0.984 & 0.996 & 0.997 & 0.9875   \\ \hline
IMOG   & 0.984 & 0.995 & 0.995 & 0.999 & 0.99325  \\ \hline
LBPH   & 0.991 & 0.987 & 0.999 & 0.996 & 0.999325 \\ \hline
STBS   & 0.992 & 0.993 & 0.99  & 0.998 & 0.99325  \\ \hline
OCTREE & 0.992 & 0.999 & 0.999 & 0.991 & 0.99525  \\ \hline
\end{tabular}\\ 
\caption{Comparision of $F_0$ scores on different datasets}
\label{my-label}
\end{table}


%


%

%

\end{document}